\documentclass[11pt]{article}
\newlength{\defbaselineskip}
\usepackage{algorithm}
\usepackage{algpseudocode}
\usepackage{framed}
\usepackage{amssymb}
\usepackage{amsfonts}
\usepackage{amsmath,amsthm}
\usepackage{graphicx}
\usepackage{caption}
\usepackage{subcaption}
\usepackage{url}
\usepackage{rotating}
\usepackage{multirow}
\usepackage{color}
\usepackage{xcolor}
\usepackage{enumitem}
\usepackage{epstopdf}
\usepackage{hyperref}
\def\BC{\begin{center}}
\def\EC{\end{center}}
\def\BIT{\begin{itemize}}
\def\EIT{\end{itemize}}
\def\BET{\begin{enumerate}}
\def\EET{\end{enumerate}}
\def\BEQ{\begin{equation}}
\def\EEQ{\end{equation}}
\newcommand{\R}{\mathbb{R}}
\newcommand{\prob}{\mathsf{Pr}}
\newcommand{\risk}{\mathcal{R}}
\newcommand{\ex}{\mathsf{E}}
\newcommand{\F}{\mathcal{F}}
\newcommand{\X}{\mathcal{X}}
\newcommand{\Y}{\mathcal{Y}}
\newcommand{\trace}{\mathsf{Tr}}
\newcommand{\diag}{\textup{diag}}

\newcommand{\lbr}{\langle}
\newcommand{\rbr}{\rangle}

\newtheorem{thm}{Theorem}
\newtheorem{definition}{Definition}

\newtheorem{lemma}{Lemma}

\newenvironment{proofof}[1]{{\em Proof of #1.}}{\hfill%\rule{2mm}{2mm}
\qed}
\begin{document}

\title{
Fast Randomized Kernel Methods With Statistical Guarantees
}

\author{
Ahmed {El Alaoui}
\thanks{
Department of Electrical Engineering and Computer Sciences,
University of California at Berkeley,
Berkeley, CA 94720.
Email:  elalaoui@eecs.berkeley.edu
}
\and
Michael W. Mahoney
\thanks{
International Computer Science Institute 
and Department of Statistics,
University of California at Berkeley,
Berkeley, CA 94720.
Email:  mmahoney@stat.berkeley.edu
}
}

%\date{\today}
\date{}

\maketitle

%----------------------------------------------------------------------- 
% Main text
%----------------------------------------------------------------------- 

\begin{abstract}
One approach to improving the running time of kernel-based methods is to 
build a small sketch of the kernel matrix and use it in lieu of the full 
matrix in the machine learning task of interest. 
Here, we describe a version of this approach that comes with running time 
guarantees as well as improved guarantees on its statistical performance.
By extending the notion of \emph{statistical leverage scores} to the setting 
of kernel ridge regression, we are able to identify a sampling distribution 
that reduces the size of the sketch (i.e., the required number of columns to 
be sampled) to the \emph{effective dimensionality} of the problem.  
This latter quantity is often much smaller than previous bounds that depend on the 
\emph{maximal degrees of freedom}. We give an empirical evidence supporting this fact. 
Our second contribution is to present a fast algorithm to quickly compute coarse approximations to these
scores in time linear in the number of samples. More precisely, the running time of the algorithm is $O(np^2)$ with $p$ only depending on the trace of the kernel matrix and the regularization parameter. This is obtained via a variant of squared length sampling that we adapt to the kernel setting. Lastly, we discuss how this new notion of the leverage of a data point captures a fine notion of the difficulty of the learning problem. 
\end{abstract}

%=====================================================================
%
%	Introduction
%
%=====================================================================
%\vspace{-.5cm}
\section{Introduction}
We consider the low-rank approximation of symmetric positive semi-definite 
(SPSD) matrices that arise in machine learning and data analysis, with an 
emphasis on obtaining good statistical guarantees.
This is of interest primarily in connection with kernel-based machine 
learning methods.
Recent work in this area has focused on one or the other of two very 
different perspectives: an \emph{algorithmic perspective}, where the focus 
is on running time issues and worst-case quality-of-approximation 
guarantees, given a fixed input matrix; and a \emph{statistical 
perspective}, where the goal is to obtain good inferential properties, under 
some hypothesized model, by using the low-rank approximation in place of the 
full kernel matrix.
The recent 
%%empirical and theoretical 
results of Gittens and 
Mahoney~\cite{gittens2013revisiting} 
%%(see also references therein) 
provide the strongest 
example of the former, and the recent results of Bach~\cite{bach2012sharp} are an 
excellent example of the latter.
In this paper, we combine ideas from these two lines of work in order to 
obtain a fast randomized kernel method with statistical guarantees that
are improved relative to the state-of-the-art.

To understand our approach, recall that several papers have established the 
crucial importance---from the algorithmic perspective---of the 
\emph{statistical leverage scores}, as they capture structural non-uniformities of the input 
matrix and they can be used to obtain very sharp worst-case approximation 
guarantees.
See, e.g., work on CUR matrix 
decompositions~\cite{drineas2008relative,mahoney2009cur}, work on 
the the fast approximation of the statistical leverage scores~\cite{drineas2012fast}, 
and the recent review~\cite{mahoney2011randomized} for more details.
%%(More generally than just SPSD matrices, randomized matrix algorithms that 
%%construct such low-rank approximations by sampling a small number of 
%%columns~\cite{drineas2006fast2,drineas2008relative,frieze2004fast} or elements~\cite{achlioptas2001fast,recht2011simpler} 
%%from the original matrix, or that perform random projections, have become 
%%widely popular.)
Here, we simply note that, when restricted to an $n \times n$ SPSD 
matrix $K$ and a rank parameter $k$, the statistical leverage scores 
relative to the best rank-$k$ approximation to $K$, call them $\ell_i$, for 
$i\in\{1,\ldots,n\}$, are the diagonal elements of the projection matrix 
onto the best rank-$k$ approximation of $K$.
That is, $\ell_i = \text{diag}(K_kK_k^{\dagger})_i$, where $K_k$ is the best rank $k$ 
approximation of $K$ and where $K_k^{\dagger}$ is the Moore-Penrose inverse 
of $K_k$.
The recent work by Gittens and Mahoney \cite{gittens2013revisiting} showed that 
qualitatively improved worst-case bounds for the low-rank approximation of 
SPSD matrices could be obtained in one of two related ways: 
either compute (with the fast algorithm of~\cite{drineas2012fast}) approximations to 
the leverage scores, and use those approximations as an importance sampling 
distribution in a random sampling algorithm; or 
rotate (with a Gaussian-based or Hadamard-based random projection) to a 
random basis where those scores are uniformized, and sample randomly in that 
rotated basis.

In this paper, we extend these ideas, and we show that---from the statistical 
perspective---we are able to obtain a low-rank approximation that comes with 
improved statistical guarantees by using a variant of this more traditional 
notion of statistical leverage. 
In particular, we improve the recent bounds of Bach~\cite{bach2012sharp}, which 
provides the first known statistical convergence result when substituting 
the kernel matrix by its low-rank approximation. 
To understand the connection, recall that a key component of Bach's approach 
is the quantity $d_{\text{mof}}=n\|\,\diag(\,K(K+n\lambda I)^{-1})\|_{\infty}$, which he 
calls the \emph{maximal marginal degrees of freedom}.\footnote{We will refer to it as the maximal degrees of freedom.} 
Bach's main result is that by constructing a low-rank approximation of the 
original kernel matrix by sampling uniformly at random $p=O(d_{\text{mof}}/\epsilon)$ 
columns, i.e., performing the vanilla Nystr\"{o}m method, and then by using 
this low-rank approximation in a prediction task, the statistical 
performance is within a factor of $1+ \epsilon$ of the performance when the 
entire kernel matrix is used. 
Here, we show that this uniform sampling is suboptimal.
We do so by sampling with respect to a coarse but quickly-computable 
approximation of a variant to the statistical leverage scores, given in 
Definition~\ref{lev} below, and we show that we can obtain similar 
$1+\epsilon$ guarantees by sampling only $O(d_{\text{eff}}/\epsilon)$ 
columns, where $d_{\text{eff}}= \trace(K(K+n\lambda I)^{-1}) < d_{\text{mof}}$.
The quantity $d_{\text{eff}}$ is called the \emph{effective dimensionality} 
of the learning problem, and it can be interpreted as the implicit number of 
parameters in this nonparametric setting \cite{zhang2013divide, friedman2001elements}. 

We expect that our results and insights will be useful much more generally.
As an example of this, we can directly compare the Nystr\"{o}m sampling 
method to a related divide-and-conquer approach, thereby answering an open 
problem of Zhang et al.\ \cite{zhang2013divide}. 
Recall that the Zhang et al.\ divide-and-conquer method consists of dividing 
the dataset $\{(x_i,y_i)\}_{i=1}^n$ into $m$ random partitions of equal 
size, computing estimators on each partition in parallel, and then averaging 
the estimators. 
They prove the minimax optimality of their estimator, although their multiplicative constants are suboptimal; and, in terms of the number of kernel 
evaluations, their method requires $m\times (n/m)^2$, with $m$ in the order 
of $n/d_{\text{eff}}^2$, which gives a total number of 
$O(nd_{\text{eff}}^2)$ evaluations. 
They noticed that the scaling of their estimator was \emph{not} directly 
comparable to that of the Nystr\"{o}m sampling method (which was proven to only require $O(nd_{\text{mof}})$ evaluations, if the sampling is 
uniform \cite{bach2012sharp}), and they left it as an open problem  to determine
which if either method is fundamentally better than the other. 
Using our Theorem \ref{main}, we are able to put both results on a common ground for 
comparison. 
Indeed, the estimator obtained by our \emph{non-uniform} Nystr\"{o}m 
sampling requires only $O(n d_{\text{eff}})$ kernel evaluations (compared to 
$O(nd_{\text{eff}}^2)$ and $O(nd_{\text{mof}})$), and it obtains the same bound on the 
statistical predictive performance as in \cite{bach2012sharp}. 
In this sense, our result combines ``the best of both worlds,'' by having 
the reduced sample complexity of \cite{zhang2013divide} and the sharp approximation 
bound of \cite{bach2012sharp}.

%
%=====================================================================
%
%	Preliminaries
%
%=====================================================================
%
%\vspace{-.3cm}
\section{Preliminaries and notation}
Let $\{(x_i,y_i)\}_{i = 1}^n$ be $n$ pairs of points in 
$\X \times \Y$, where $\X$ is the input space and $\Y$ is the response space. 
The kernel-based learning problem can be cast as the following minimization 
problem: 
\BEQ
\min_{f\in \F} \frac{1}{n}\sum_{i=1}^n\ell(y_i,f(x_i)) + \frac{\lambda}{2} \|f\|_\F^2  ,
\label{ker}
\EEQ
where $\F$ is a reproducing kernel Hilbert space and 
$\ell:\Y\times\Y \rightarrow \R$ is a loss function. 
We denote by $k:\X\times\X \rightarrow \R$ the positive definite kernel 
corresponding to $\F$ and by $\phi: \X \rightarrow \F$ a corresponding 
feature map. 
That is, $k(x,x') = \lbr \phi(x),\phi(x')\rbr_\F$ for every $x,x' \in \X$. 
The representer theorem \cite{kimeldorf1971some, scholkopf2001generalized} allows us to reduce 
Problem~\eqref{ker} to a finite-dimensional optimization problem, 
in which case 
%%Indeed, it is guaranteed that the solution to \eqref{ker} is of the form 
%%$\hat{f} = \sum_{i=1}^n \alpha_i k(x_i,\cdot)$. 
Problem~\eqref{ker} boils down to finding the vector $\alpha \in \R^n$ 
that solves 
\BEQ
 \min_{\alpha \in \R^n} \frac{1}{n}\sum_{i=1}^n\ell(y_i,(K\alpha)_i) + \frac{\lambda}{2} \alpha^\top K\alpha,
 \label{dis_ker}
\EEQ
where $K_{ij} = k(x_i,x_j)$. 
We let $U\Sigma U^\top $ be the eigenvalue decomposition of $K$, with 
$\Sigma = \text{Diag}(\sigma_1,\cdots,\sigma_n)$, $\sigma_1 \geq\cdots \geq \sigma_n\geq 0$,  
and $U$ an orthogonal matrix. The underlying data model is 
\[y_i = f^*(x_i) + \sigma^2 \xi_i ~~~ i = 1, \cdots,n\] 
with $f^* \in \F$, $(x_i)_{1\leq i\leq n}$ a deterministic sequence and $\xi_i$ are i.i.d.\ standard normal random 
variables. We consider $\ell$ to be the squared loss, in which case we will be interested in the mean squared error as a measure of statistical risk: for any estimator $\hat{f}$, let 
\begin{equation}
\risk(\hat{f}) := \frac{1}{n}\ex_{\xi} \|\hat{f} - f^*\|_2^2
\label{risk}
\end{equation} 
be the risk function of $\hat{f}$ where $\ex_{\xi}$ denotes the expectation under the randomness induced by $\xi$.
In this setting the problem is called 
\emph{Kernel Ridge Regression} (KRR).
The solution to Problem~\eqref{dis_ker} is $\alpha = (K+n\lambda I)^{-1}y$, 
and the estimate of $f^*$ at any training point $x_i$ is given by 
$\hat{f}(x_i) = (K\alpha)_i$. 
We will use $\hat{f}_K$ as a shorthand for 
the vector $(\hat{f}(x_i))_{1\leq i\leq n} \in \R^n$ when the matrix $K$ is used as 
a kernel matrix. This notation will be used accordingly for other kernel matrices (e.g.\ $\hat{f}_L$ for a matrix $L$). Recall that the risk of the estimator $\hat{f}_K$ can then be decomposed into a bias and variance term:
\begin{align}
\mathcal{R}(\hat{f}_K)  &= \frac{1}{n}\ex_{\xi} \|K(K+n\lambda I)^{-1}(f^*+\sigma^2\xi) - f^*\|_2^2  \nonumber\\ 
%= \frac{1}{n}\ex_{\xi} \|\hat{f}_K - f^*\|_2^2
&= \frac{1}{n}\|(K(K+n\lambda I)^{-1}-I)f^*\|_2^2 + \frac{\sigma^2}{n}\ex_{\xi} \|K(K+n\lambda I)^{-1}\xi\|_2^2\nonumber \\
&= n\lambda^2 \|(K+n\lambda I)^{-1}f^*\|_2^2 + \frac{\sigma^2}{n}\trace(K^2(K+n\lambda I)^{-2}) \nonumber\\
&:= ~~~~~~~~~ \text{bias}(K)^2 ~~~~~~~~ + ~~~~~~~ \text{variance}(K). \label{decomp}
\end{align}

%The expressions above suggest that the variance should be matrix-increasing and the bias should be matrix-decreasing in $K$. It turns out that the first claim is true while the second is not in general.  
Solving Problem~\eqref{dis_ker}, either by a direct method or by an 
optimization algorithm needs at least a quadratic and often cubic running 
time in $n$ which is prohibitive in the large scale setting. 
The so-called Nytr\"{o}m method approximates the solution to 
Problem~\eqref{dis_ker} by substituting $K$ with a low-rank approximation 
to $K$. 
In practice, this approximation is often not only fast to construct, but the 
resulting learning problem is also often easier to 
solve \cite{fine2002efficient,williams2001using,kumar2009sampling,gittens2013revisiting}. 
The method operates as follows. 
A small number of columns $K_1,\cdots,K_p$ are randomly sampled from $K$. 
If we let $C = [K_1,\cdots,K_p] \in \R^{n \times p}$ denote the matrix 
containing the sampled columns, $W \in \R^{p \times p}$ the overlap between 
$C$ and $C^\top$ in $K$, then the Nystr\"{o}m approximation of $K$ is the 
matrix 
\[L = CW^{\dagger}C^\top .\]     
%$ L = CW^{\dagger}C^\top $.
More generally, if we let $S \in R^{n\times p}$ be an arbitrary 
\emph{sketching matrix}, i.e., a tall and skinny matrix that, when 
left-multiplied by $K$, produces a ``sketch'' of $K$ that preserves some 
desirable properties, then the Nystr\"{o}m approximation associated with $S$ is 
%$L = KS(S^\top KS)^{\dagger}S^\top K $. 
\[L = KS(S^\top KS)^{\dagger}S^\top K.\] 
For instance, for random sampling algorithms, $S$ would contain a non-zero 
entry at position $(i,j)$ if the $i$-th column of $K$ is chosen at the $j$-th trial of 
the sampling process.
Alternatively, $S$ could also be a random projection matrix; or $S$ could 
be constructed with some other (perhaps deterministic) method, as long as 
it verifies some structural properties, depending on the 
application~\cite{mahoney2011randomized,gittens2013revisiting,mahoney2009cur,drineas2008relative}.

We will focus in this paper on analyzing this approximation in the 
statistical prediction context related to the estimation of $f^*$ by 
solving Problem~\eqref{dis_ker}. 
We proceed by revisiting and improving upon prior results from three different areas. The first result (Theorem \ref{bias_}) is on the behavior of the bias of $\hat{f}_L$, when $L$ is constructed using a general sketching matrix $S$. This result underlies the statistical analysis of the Nystr\"{o}m method. To see this, first, it is not hard to prove that $L \preceq K$ in the sense of usual the order on the positive semi-definite cone. Second, one can prove that the variance is matrix-increasing, hence the variance will decrease when replacing $K$ by $L$. On the other hand, the bias (while \emph{not} matrix monotone in general) can be proven to not increase too much when replacing $K$ by $L$. This latter statement will be the main technical difficulty for obtaining a bound on $\risk(\hat{f}_L)$ (see Appendix A). A form of this result is due to Bach \cite{bach2012sharp} in the case where $S$ is a uniform sampling matrix. The second result (Theorem \ref{concent}) is a concentration bound for approximating matrix multiplication when the rank-one components of the product are sampled non uniformly. This result is derived from the matrix Bernstein inequality, and yields a sharp quantification of the deviation of the approximation from the true product. The third result (Definition \ref{lev}) is an extension of the definition of the leverage scores to the context of kernel ridge regression. Whereas the notion of leverage is established as an algorithmic tool in randomized linear algebra, we introduce a natural counterpart of it to this statistical setting. By combining these contributions, we are able to give a sharp statistical statement on the behavior of the Nystr\"{o}m method if one is allowed to sample non uniformly. All the proofs are deferred to the appendix. 
% NOTE: eq nb. in appendix A : \eqref{order} to be placed in paragraph above in final submission
 %when $S$ is a sampling matrix with a general sampling distribution, although  

%i.e. the elements of $W$ appear in $C$ and in $C^\top$ ($W$ is always non empty if $p\geq 1$ and $K$ is symmetric)

%
%=====================================================================
%
%	Revisiting prior work and new results
%
%=====================================================================
%
%\vspace{-.3cm}
\section{Revisiting prior work and new results}
\subsection{A structural result} 
We begin by stating a ``structural" result that upper-bounds the bias of the estimator constructed using the approximation $L$. This result is deterministic: it only depends on the properties of the input data, and holds for \emph{any} sketching matrix $S$ that satisfies certain conditions. This way the randomness of the construction of $S$ is decoupled from the rest of the analysis. We highlight the fact that this view offers a possible way of improving the current results since a better construction of $S$ -whether deterministic or random- satisfying the data-related conditions would immediately lead to down stream algorithmic and statistical improvements in this setting.     

\begin{thm}
Let $S \in \R^{n \times p}$ be a sketching matrix and $L$ the corresponding Nystr\"{o}m approximation. %and $L_\gamma$ a regularized version of it: $L_\gamma = KS(S^\top KS + \gamma I)^{-1}S^\top K$ for $\gamma >0$. 
%Denote by $\text{bias}(K)$ the bias of the KRR estimator $\hat{f}_K$ with regularization parameter $\lambda >0$ as in the previous section. 
For $\gamma > 0$, let $\Phi = \Sigma(\Sigma +n\gamma I)^{-1}$. If the sketching matrix $S$ satisfies $\lambda_{\max}\Big(\Phi - \Phi^{1/2} U^\top SS^\top U\Phi^{1/2} \Big)\leq t$ for $t \in (0,1)$ and $\lambda \ge \frac{1}{1-t}\|S\|_{\text{op}}^2 \cdot\frac{\lambda_{\max}(K)}{n}$,  where $\lambda_{\max}$ denotes the maximum eigenvalue and $\|\cdot\|_{\text{op}}$ is the operator norm then %and $\gamma \leq (1-t) \lambda$
\begin{equation}
\text{bias}(L) \leq \left(1+ \frac{\gamma/\lambda}{1-t}\right) \text{bias}(K).
\label{bias}
\end{equation}
\label{bias_}
\end{thm}
In the special case where $S$ contains one non zero entry equal to $1/\sqrt{pn}$ in every column with $p$ the number of sampled columns, the result and its proof can be found in \cite{bach2012sharp} (appendix B.2), although we believe that their argument contains a problematic statement. We propose an alternative and complete proof in Appendix A. The subsequent analysis unfolds in two steps: (1) assuming the sketching matrix $S$ satisfies the conditions stated in Theorem \ref{bias_}, we will have $\risk(\hat{f}_L) \lesssim \risk(\hat{f}_K) $, and (2) matrix concentration is used to show that an appropriate random construction of $S$ satisfies the said conditions. We start by stating the concentration result that is the source of our improvement (section 3.2), define a notion of statistical leverage scores (section 3.3), and then state and prove the main statistical result (Theorem \ref{main} section 3.4). We then present our main algorithmic result consisting of a fast approximation to this new notion of leverage scores (section 3.5).      

\subsection{A concentration bound on matrix multiplication}
Next, we state our result for approximating matrix products of the form $\Psi\Psi^\top$ when a few columns from $\Psi$ are sampled to form the approximate product $\Psi_I\Psi_I^\top$ where $\Psi_I$ contains the chosen columns. The proof relies on a matrix Bernstein inequality (see e.g. \cite{tropp2012user}) and is presented at the end of the paper (Appendix B).  
\begin{thm}
Let $n, m$ be positive integers. Consider a matrix $\Psi \in \R^{n\times m}$ and denote by $\psi_{i}$ the $i^{th}$ column of $\Psi$. Let $p \leq m$ and $I=\{i_1,\cdots,i_p\}$ be a subset of $\{1,\cdots, m\}$ formed by $p$ elements chosen randomly with replacement, according to the distribution 
\begin{equation}
 \forall i \in \{1,\cdots,m\} \quad \prob(\text{choosing } i)  = p_i \geq \beta  \frac{\|\psi_i\|_2^2}{\|\Psi\|_F^2} \label{robust}
 \end{equation} 
 for some $\beta \in (0,1]$. %Let $\Psi_I$ denote the $n \times p$ matrix consisting of the subset of columns   $P = p\cdot\text{diag}(p_{i_1},\cdots,p_{i_p})$ 
Let $S\in \R^{n\times p}$ be a sketching matrix such that $S_{ij} = 1/\sqrt{p\cdot p_{i_j}}$ only if $i=i_j$ and 0 elsewhere. Then 
\begin{equation}
\prob  \Big(\lambda_{\max}\big(\Psi \Psi^\top - \Psi SS^\top\Psi^\top\big) \geq t\Big) 
\leq n\exp\left(\frac{-pt^2/2}{\lambda_{\max}(\Psi \Psi^\top)(\|\Psi\|_F^2/\beta + t/3)}\right). \label{dev}
\end{equation}
\label{concent}
\end{thm}
\textbf{Remarks:} 1. This result will be used for  $\Psi = \Phi^{1/2}U^\top$, in conjunction with Theorem \ref{bias_} to prove our main result in Theorem \ref{main}. Notice that $\Psi^\top$ is a scaled version of the eigenvectors, with a scaling given by the diagonal matrix $\Phi = \Sigma(\Sigma +n\gamma I)^{-1}$ which should be considered as ``soft projection" matrix that smoothly selects the top part of the spectrum of $K$. The setting of Gittens et al.\ \cite{gittens2013revisiting}, in which $\Phi$ is a 0-1 diagonal is the closest analog of our setting.
   
2. It is known that $p_i=\frac{\|\psi_i\|_2^2}{\|\Psi\|_F^2}$ is the optimal sampling distribution in terms of minimizing the expected error $\ex \|\Psi\Psi^\top - \Psi SS^\top\Psi^\top\|_F^2$ \cite{drineas2006fast1}. The above result exhibits a robustness property by allowing the chosen sampling distribution to be different from the optimal one by a factor $\beta$.\footnote{In their work \cite{drineas2006fast1}, Drineas et al.\ have a comparable robust statement for controlling the expected error. Our result is a robust quantification of the tail probability of the error, which is a much stronger statement.} The sub-optimality of such a distribution is reflected in the upper bound (\ref{dev}) by the amplification of the squared Frobenius norm of $\Psi$ by a factor $1/\beta$. For instance, if the sampling distribution is chosen to be uniform, i.e. $p_i = 1/m$, then the value of $\beta$ for which (\ref{robust}) is tight is $\frac{\|\Psi\|_F^2}{m\max_i \|\psi_i\|_2^2},$ in which case we recover a concentration result proven by Bach \cite{bach2012sharp}. Note that Theorem \ref{concent} is derived from one of the state-of-the-art bounds on matrix concentration, but it is one among many others in the literature; and while it constitutes the base of our improvement, it is possible that a concentration bound more tailored to the problem might yield sharper results.  

\subsection{An extended definition of leverage}
We introduce an extended notion of leverage scores that is specifically tailored to the ridge regression problem, and that we call the \emph{$\lambda$-ridge leverage scores}.

\begin{definition}
For $\lambda >0$, the $\lambda$-ridge leverage scores associated with the kernel matrix $K$ and the parameter $\lambda$ are
\begin{equation}
\forall i \in \{1,\cdots,n\}, ~~~ \quad l_i(\lambda) = \sum_{j=1}^n \frac{\sigma_j}{\sigma_j + n\lambda}U_{ij}^2. 
\end{equation}
\label{lev}
\end{definition}

Note that $l_i(\lambda)$ is the $i^{th}$ diagonal entry of $K(K+n\lambda I)^{-1}$. %Moreover, for $\Psi = \Sigma^{1/2}(\Sigma +\lambda I)^{-1/2}U^\top$, we have $\|\psi_i\|_2^2 = l_i(\lambda)$. This matrix will appear later in the analysis of our main result. 
The quantities $(l_i(\lambda))_{1\leq i\leq n}$ are in this setting the analogs of the so-called \emph{leverage scores} in the statistical literature, as they characterize the data points that ``stick out", and consequently that most affect the result of a statistical procedure. They are classically defined as the row norms of the left singular matrix $U$ of the input matrix, and they have been used in regression diagnostics for outlier detection \cite{chatterjee1986influential}, and more recently in randomized matrix algorithms as they often provide an optimal importance sampling distribution for  constructing random sketches for low rank approximation \cite{drineas2006fast1, drineas2006fast2, drineas2008relative, mahoney2009cur, gittens2013revisiting} and least squares regression \cite{drineas2011faster} when the input matrix is tall and skinny ($n \geq m$).  In the case where the input matrix is square, this definition is vacuous as the row norms of $U$ are all equal to 1. Recently, Gittens and Mahoney \cite{gittens2013revisiting} used a truncated version of these scores (that they called \emph{leverage scores relative to the best rank-$k$ space}) to obtain the best algorithmic results known to date on low rank approximation of positive semi-definite matrices. Definition \ref{lev} is a weighted version of the classical leverage scores, where the weights depend on the spectrum of $K$ and a regularization parameter $\lambda$. In this sense, it is an interpolation between Gittens' scores and the classical (tall-and-skinny) leverage scores, where the parameter $\lambda$  plays the role of a rank parameter. In addition, we point out that Bach's maximal degrees of freedom $d_{\text{mof}}$ is to the $\lambda$-ridge leverage scores what the \emph{coherence} is to Gittens' leverage scores, i.e. their (scaled) maximum value: $d_{\text{mof}}/n=\max_i l_i(\lambda)$; and that while the sum of Gittens' scores is the rank parameter $k$, the sum of the $\lambda$-ridge leverage scores is the effective dimensionality $d_{\text{eff}}$. We argue in the following that Definition \ref{lev} provides a relevant notion of leverage in the context of kernel ridge regression. It is the natural counterpart of the algorithmic notion of leverage in the prediction context. We use it in the next section to make a statistical statement on the performance of the Nystr\"{o}m method.

\subsection{Main statistical result: an error bound on approximate kernel ridge regression}
Now we are able to give an improved version of a theorem by Bach \cite{bach2012sharp} that establishes a performance guaranty on the use of the Nystr\"{o}m method in the context of kernel ridge regression. It is improved in the sense that the sufficient number of columns that should be sampled in order to incur no (or little) loss in the prediction performance is lower. This is due to a more data-sensitive way of sampling the columns of $K$ (depending on the $\lambda$-ridge leverage scores) during the construction of the approximation $L$. The proof is in Appendix C.   

\begin{thm}
Let $\lambda, \epsilon >0$, $\rho \in (0,1/2)$, $n \ge 2$ and $L$ be a Nystr\"{o}m approximation of $K$ by choosing $p$ columns randomly with replacement according to a probability distribution $(p_i)_{1\leq i\leq n}$ such that $\forall i \in \{1,\cdots,n\}, ~~~ p_i \geq \beta \cdot l_i(\lambda\epsilon)/\sum_{i=1}^n l_i(\lambda\epsilon)$ for some $\beta \in (0,1]$. Let $\underline{l} \le \min_i l_i(\lambda \epsilon)$. 
If \[p \geq 8\left(\frac{d_{\text{eff}}}{\beta} + \frac{1}{6}\right)\log \left(\frac{n}{\rho}\right) ~~\text{and}~~\lambda \ge 2\left(1+\frac{1}{\underline{l}}\right) \frac{\lambda_{\max}(K)}{n},\]
with $d_{\text{eff}} = \sum_{i=1}^n l_i(\lambda\epsilon) = \trace(K(K+n\lambda \epsilon I)^{-1})$ then 
\[\risk(\hat{f}_L)   \leq (1+2\epsilon)^{2}\risk(\hat{f}_K)  \]
with probability at least $1-2\rho$, where $(l_i)_i$ are introduced in Definition \ref{lev} and $\risk$ is defined in \eqref{risk}.
\label{main}

\end{thm}
%If \[p \geq \max\left\{8\left(\frac{d_{\text{eff}}}{\beta} + \frac{1}{6}\right)\log \left(\frac{n}{\rho}\right)~,~ \frac{8(1-\alpha)}{3\alpha}\log\left(\frac{2n}{\rho}\right)\right\} ~~\text{and}~~\lambda \ge 4 \frac{\lambda_{\max}(K)}{n},\]
%$\lambda$ has to roughly be larger than $\lambda_{\max}(K)/n$, ignoring the term %The dependence on $1/\underline{l}$ in the bound on $\lambda$ 

Theorem \ref{main} asserts that substituting the kernel matrix $K$ by a Nystr\"{o}m approximation of rank $p$ in the KRR problem induces an arbitrarily small prediction loss, provided that $p$ scales linearly with the effective dimensionality $d_{\text{eff}}$\footnote{Note that $d_{\text{eff}}$ depends on the precision parameter $\epsilon$, which is absent in the classical definition of the effective dimensionality \cite{friedman2001elements, zhang2013divide, bach2012sharp} However, the following bound holds: $d_{\text{eff}} \leq \frac{1}{\epsilon}\trace(K(K+n\lambda I)^{-1})$.} and that $\lambda$ is not too small\footnote{This condition on $\lambda$ is not necessary if one constructs $L$ as $ KS(S^\top KS+n\lambda\epsilon I)^{-1}S^\top K$ (see proof).}. The leverage-based sampling appears to be crucial for obtaining this dependence, as the $\lambda$-ridge leverage scores provide information on which columns -and hence which data points- capture most of the difficulty of the estimation problem. Also, as a sanity check, the smaller the target accuracy $\epsilon$, the higher $d_{\text{eff}}$, and the more uniform the sampling distribution $(l_i(\lambda\epsilon))_i$ becomes. In the limit $\epsilon \rightarrow 0$, $p$ is in the order of $n$ and the scores are uniform, and the method is essentially equivalent to using the entire matrix $K$.   
Moreover, if the sampling distribution $(p_i)_i$ is a factor $\beta$ away from optimal, a slight oversampling (i.e. increase $p$ by $1/\beta$) achieves the same performance. In this sense, the above result shows robustness to the sampling distribution. This property is very beneficial from an implementation point of view, as the error bounds still hold when only an approximation of the leverage scores is available. If the columns are sampled uniformly, a worse lower bound on $p$ that depends on $d_{\text{mof}}$ is obtained \cite{bach2012sharp}.     

\subsection{Main algorithmic result: a fast approximation to the $\lambda$-ridge leverage scores}
Although the $\lambda$-ridge leverage scores can be naively computed using SVD, the exact computation is as costly as solving the original Problem~\eqref{dis_ker}. Therefore, the central role they play in the above result motivates the problem of a fast approximation, in a similar way the importance of the usual leverage scores has motivated Drineas et al.\ to approximate them is random projection time \cite{drineas2012fast}. A success in this task will allow us to combine the running time benefits with the improved statistical guarantees we have provided. %The approximation scheme we use is based on the same Nystr\"{o}m sampling ideas that were used to approximate kernel ridge regression.
%And similarly to the usual leverage scores, where their demonstrated importance has motivated Drineas et al.\ to approximate them is random projection time \cite{drineas2012fast}, the above results motivate the problem of a fast approximation of our variant of the leverage scores, thereby combining the running time benefits with the improved statistical guarantees we have provided. 
%\vspace{-.2cm}
\paragraph{Algorithm:}
 \BIT
% \item \textbf{Inputs}: $K\in \R^{n\times n}$, $\lambda >0$, $\epsilon \in (0,1/2)$, $p \in \{1,2,\cdots\}$
\item \textbf{Inputs}: data points $(x_i)_{1\leq i\leq n}$, probability vector $(p_i)_{1\leq i\leq n}$, sampling parameter $p \in \{1,2,\cdots\}$, $\lambda >0$, $\epsilon \in (0,1/2)$.
 \item \textbf{Output}: $(\tilde{l}_i)_{1\leq i \leq n}$ $\epsilon$-approximations to $(l_i(\lambda))_{1\leq i \leq n}$.
 \item[1.] Sample $p$ data points from $(x_i)_{1\leq i\leq n}$ with replacement with probabilities $(p_i)_{1\leq i\leq n}$.
 \item[2.] Compute the corresponding columns $K_1,\cdots,K_p$ of the kernel matrix. 
% \item[1.] Sample $p = p(K,\epsilon,\lambda)$ columns from $K$ with replacement with probabilities $K_{ii}/\trace(K)$.
 \item[3.] Construct $C = [K_1,\cdots,K_p] \in \R^{n\times p}$ and $W \in \R^{p\times p}$ as presented in Section 2.
 \item[4.] Construct  $B\in \R^{n\times p}$ such that $ BB^\top =  CW^\dagger C^\top$.
 \item[5.] For every $i \in \{1,\cdots,n\}$, set 
 \begin{equation} \tilde{l}_i = B_i^\top(B^\top B+n\lambda I)^{-1}B_i \label{inv} \end{equation}
 where $B_i$ is the $i$-th row of $B$, and return it. 
 \EIT
 \paragraph{Running time:} The running time of the above algorithm is dominated by steps 4 and 5. Indeed, constructing $B$ can be done using a Cholesky factorization on $W$ and then a multiplication of $C$ by the inverse of the obtained Cholesky factor, which yields a running time of $O(p^3 + np^2)$. Computing the approximate leverage scores $(\tilde{l}_i)_{1\leq i \leq n}$ in step 5 also runs in $O(p^3+np^2)$. Thus, for $p \ll n$, the overall algorithm runs in $O(np^2)$. Note that formula \eqref{inv} only involves matrices and vectors of size $p$ (everything is computed in the smaller dimension $p$), and the fact that this yields a correct approximation relies on the matrix inversion lemma (see proof in Appendix D). Also, only the relevant columns of $K$ are computed and we never have to form the entire kernel matrix. This improves over earlier models \cite{gittens2013revisiting} that require that all of $K$ has to be written down in memory. The improved running time is obtained by considering the construction \eqref{inv} which is quite different from the regular setting of approximating the leverage scores of a rectangular matrix \cite{drineas2012fast}. We now give both additive and multiplicative error bounds on its approximation quality.  
\begin{thm}
Let $\epsilon \in (0,1/2)$, $\rho \in (0,1)$ and $\lambda >0$. Let $L$ be a Nystr\"{o}m approximation of $K$ by choosing $p$  columns at random with probabilities $p_i = K_{ii}/\trace(K)$, $i=1,\cdots,n$. If\[p\geq 8\left(\frac{\trace(K)}{n\lambda\epsilon}+ \frac{1}{6}\right)\log\left(\frac{n}{\rho}\right)\] then we have $\forall i \in \{1,\cdots,n\}$
\[\text{\small{(additive error bound)}} ~~~ l_i(\lambda)-2\epsilon \leq \tilde{l}_i\leq l_i(\lambda)  \]
and
\[\text{\small{(multiplicative error bound)}}  ~  \Big(\frac{\sigma_{n}-n\lambda\epsilon}{\sigma_{n}+n\lambda\epsilon}\Big)l_i(\lambda) \leq \tilde{l}_i\leq l_i(\lambda)\]
with probability at least $1-\rho$. 
\label{approx_lev}
\end{thm}
%\vspace{-.2cm}
\textbf{Remarks:} 1. Theorem \ref{approx_lev} states that if the columns of $K$ are sampled proportionally to $K_{ii}$ then $O(\frac{\trace(K)}{n\lambda})$ is a sufficient number of samples. Recall that $K_{ii} = \|\phi(x_i)\|_{\F}^2$, so our procedure is akin to sampling according to the squared lengths of the data vectors, which has been extensively used in different contexts of randomized matrix approximation \cite{frieze2004fast, drineas2006fast1, drineas2006fast2, mahoney2011randomized, gittens2013revisiting}.  

%2. In virtue of Theorem \ref{main}, the above multiplicative error bound ---although weaker than the additive error bound since it depends on the minimum eigenvalue of $K$--- directly provides a bound on the prediction error when $(\tilde{l}_i)_i$ are used as a sampling distribution.

2. Due to how $\lambda$ is defined in eq.\ \eqref{ker} the $n$ in the denominator is artificial: $n\lambda$ should be thought of as a ``rescaled" regularization parameter $\lambda'$. In some settings, the $\lambda$ that yields the best generalization error scales like $O(1/\sqrt{n})$, hence $p =O(\trace(K)/\sqrt{n})$ is sufficient. On the other hand, if the columns are sampled uniformly, one would get $p =O(d_{\text{mof}}) = O(n\max_i l_i(\lambda))$. 

\section{Experiments}
\begin{figure}[t]
%\vspace{-.2cm}
\centering
\includegraphics[width=13cm,height=5.6cm]{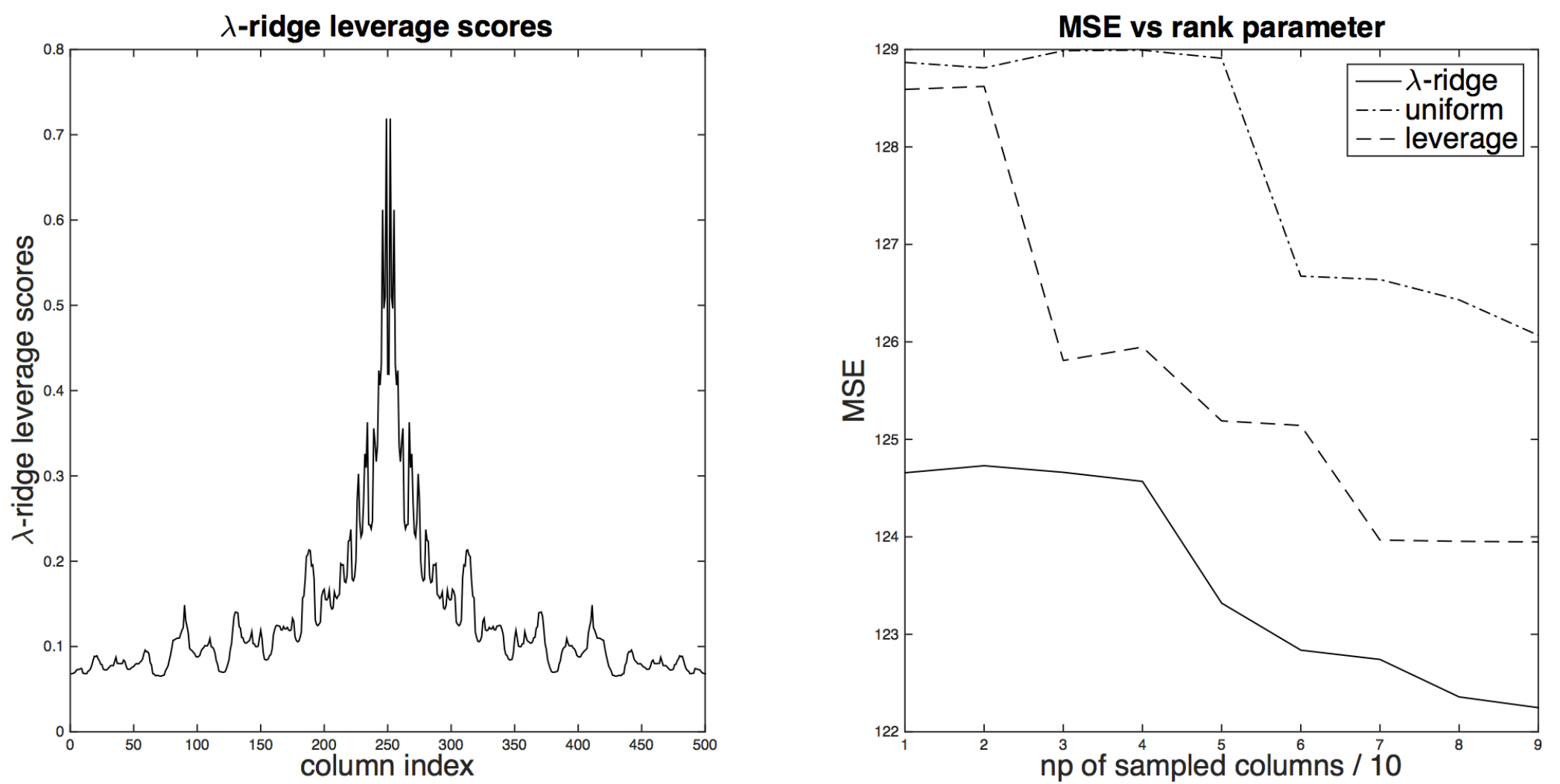}
\caption{ \small{The $\lambda$-ridge leverage scores for the synthetic Bernoulli data set described in the text (left) and the MSE risk vs.\ the number of sampled columns used to construct the Nystr\"{o}m approximation for different sampling methods (right).}} %and Gittens et al.\  leverage scores
\label{fig1}
\end{figure}

We test our results based on several datasets: one synthetic regression problem from \cite{bach2012sharp} to illustrate the importance of the $\lambda$-ridge leverage scores, the \emph{Pumadyn} family consisting of three datasets \emph{pumadyn-32fm}, \emph{pumadyn-32fh} and \emph{pumadyn-32nh} \footnote{\small{\url{http://www.cs.toronto.edu/~delve/data/pumadyn/desc.html}}} and the \emph{Gas Sensor Array Drift Dataset} from the UCI database\footnote{\small{\url{https://archive.ics.uci.edu/ml/datasets/Gas+Sensor+Array+Drift+Dataset}}}. The synthetic case consists of a regression problem on the interval $\X = [0,1]$ where, given a sequence $(x_i)_{1\leq i\leq n}$ and a sequence of noise $(\epsilon_i)_{1\leq i\leq n}$, we observe the sequence
  \[y_i = f(x_i) + \sigma^2\epsilon_i, \quad i \in \{1,\cdots,n\}.\] 
The function $f$ belongs to the RKHS $\F$ generated by the kernel $k(x,y) = \frac{1}{(2\beta)!}B_{2\beta}(x-y - \lfloor x-y\rfloor)$ where $B_{2\beta}$ is the $2\beta$-th Bernoulli polynomial \cite{bach2012sharp}. One important feature of this regression problem is the distribution of the points $(x_i)_{1\leq i\leq n}$ on the interval $\X$: if they are spread uniformly over the interval, the $\lambda$-ridge leverage scores $(l_i(\lambda))_{1\leq i\leq n}$ are uniform for every $\lambda >0$, and uniform column sampling is optimal in this case. In fact, if $x_i = \frac{i-1}{n}$ for $i=1,\cdots,n$, the kernel matrix $K$ is a circulant matrix \cite{bach2012sharp}, in which case, we can prove that the $\lambda$-ridge leverage scores are constant. 
Otherwise, if the data points are distributed asymmetrically on the interval, the $\lambda$-ridge leverage scores are non uniform, and importance sampling is beneficial (Figure \ref{fig1}). In this experiment, the data points $x_i \in (0,1)$ have been generated with a distribution symmetric about $\frac{1}{2}$, having a high density on the borders of the interval $(0,1)$ and a low density on the center of the interval. The number of observations is $n=500$. On Figure \ref{fig1}, we can see that there are few data points with high leverage, and those correspond to the region that is underrepresented in the data sample (i.e. the region close to the center of the interval since it is the one that has the lowest density of observations). The $\lambda$-ridge leverage scores are able to capture the importance of these data points, thus providing a way to detect them (e.g.\ with an analysis of outliers), had we not known their existence.   

For all datasets, we determine $\lambda$ and the band width of $k$ by cross validation, and we compute the effective dimensionality $d_{\text{eff}}$ and the maximal degrees of freedom $d_{\text{mof}}$. Table \ref{tab} summarizes the experiments. It is often the case that $d_{\text{eff}} \ll d_{\text{mof}}$ and $\risk(\hat{f}_L)/\risk(\hat{f}_K) \simeq 1$, in agreement with Theorem \ref{main}. 

\vspace{.15cm}

 \begin{table}[h]
 \centering
\resizebox{13cm}{2.5cm}{
%\hspace{-.5cm}
%\centering
 \begin{tabular}{| c | c | c | c | c | c | c | c | c |}
 \hline
kernel & dataset & $n$ & nb. feat & $\text{band width}$ & $\lambda$ & $d_{\text{eff}}$ & $d_{\text{mof}}$ & risk ratio $\risk(\hat{f}_L)/\risk(\hat{f}_K)$\\ \hline
Bern & Synth &  500 & - & - &$1e{-6}$ & 24 & 500  & 1.01 ~~ ($p=2d_{\text{eff}}$)\\ \hline
\multirow{5}{*}{Linear}
  & Gas2 & 1244 & 128 & - & $1e{-3}$& 126 & 1244  & 1.10 ~~($p=2d_{\text{eff}}$)\\ \cline{2-2}
  & Gas3 & 1586 & 128 & - & $1e{-3}$& 125& 1586 & 1.09 ~~($p=2d_{\text{eff}}$)\\ \cline{2-2}
  & Pum-32fm &  2000 & 32 & - & $1e{-3}$& 31 & 2000 & 0.99 ~~ ($p=2d_{\text{eff}}$)\\ \cline{2-2}
  & Pum-32fh & 2000 & 32 & - & $1e{-3}$& 31& 2000 & 0.99 ~~ ($p=2d_{\text{eff}}$)\\ \cline{2-2}
 & Pum-32nh &  2000 & 32 & - & $1e{-3}$& 32 & 2000 & 0.99 ~~ ($p=2d_{\text{eff}}$)\\ \hline
\multirow{5}{*}{RBF}
  & Gas2 & 1244 & - & 1 & $4.5 e{-4}$& 1135 & 1244 &1.56 ~~ ($p=d_{\text{eff}}$)\\ \cline{2-2}
  & Gas3 & 1586 & - & 1 &$5 e{-4}$ & 1450 & 1586 &1.50 ~~ ($p=d_{\text{eff}}$)\\ \cline{2-2}
  & Pum-32fm &  2000 & - & 5 & 0.5& 142& 1897 & 1.00 ~~ ($p=d_{\text{eff}}$)\\ \cline{2-2}
  & Pum-32fh & 2000 & - & 5 & $5 e{-2}$& 747& 1989 & 1.00 ~~ ($p=d_{\text{eff}}$)\\ \cline{2-2}
  & Pum-32nh &  2000 & - & 5 & $1.3 e{-2}$& 1337& 1997 & 0.99 ~~ ($p=d_{\text{eff}}$)\\ 
 \hline
\end{tabular}
}
\caption{\small{Parameters and quantities of interest for the different datasets and using different kernels: the synthetic dataset using the Bernoulli kernel (denoted by Synth), the Gas Sensor Array Drift Dataset (batches 2 and 3, denoted by Gas2 and Gas3) and the Pumadyn datasets (Pum-32fm, Pum-32fh, Pum-32nh) using linear and RBF kernels.}}
\label{tab}
\end{table}

%\vspace{-.4cm}
\section{Conclusion}
We showed in this paper that in the case of kernel ridge regression, the sampling complexity of the Nystr\"{o}m method can be reduced to the effective dimensionality of the problem, hence bridging and improving upon different previous attempts that established weaker forms of this result. This was achieved by defining a natural analog to the notion of leverage scores in this statistical context, and using it as a column sampling distribution. We obtained this result by combining and improving upon results that have emerged from two different perspectives on low rank matrix approximation. We also present a way to approximate these scores that is computationally tractable, i.e.\ runs in time $O(np^2)$ with $p$ depending only on the trace of the kernel matrix and the regularization parameter.  
One natural unanswered question is whether it is possible to further reduce the sampling complexity, or is the effective dimensionality also a lower bound on $p$? And as pointed out by previous work \cite{bach2010self,bach2012sharp}, it is likely that the same results hold for smooth losses beyond the squared loss (e.g.\ logistic regression). However the situation is unclear for non-smooth losses (e.g.\ support vector regression).  

%\vspace{-.3cm}

%\paragraph{Acknowledgements:} 

%----------------------------------------------------------------------- 
% Ack and bib
%----------------------------------------------------------------------- 

%\vspace{5mm}
\subsubsection*{Acknowledgements}
An earlier draft of this paper contained a mistake in the proof of Theorem \ref{bias_}. We thank Xixian Chen for signaling it to us. We thank Francis Bach for stimulating discussions and for contributing to a rectified proof. We thank Jason Lee and Aaditya Ramdas for fruitful discussions regarding the same issue. We thank Yuchen Zhang for pointing out the connection to his work.
\vspace{5mm}

\newpage
%\setlinespacing{1}
%\bibliographystyle{plain}
\bibliographystyle{unsrt}
\bibliography{krr}

%----------------------------------------------------------------------- 
% Appendix
%----------------------------------------------------------------------- 

\newpage
\appendix
%% MWM: The following line should be commented out if the appendix is in a different file that the main text
\section{Proof of Theorem~\ref{bias_}}
%% MWM: The following line should be commented out if the appendix is in the same file as the main text
%\section{Proof of Theorem 1}
\label{sxn:pf-bias}

\textbf{Note:} This proof is inspired by one of Bach \cite{bach2012sharp}. We extend their result to the case of a general sketching matrix $S$. Moreover, we believe their argument contains two problematic statements (about monotonicity of the bias) that we rectify with Lemma \ref{lemm2} and Lemma \ref{lemm3} below.  Their result therefore holds also true with minimal change based on this argument. %Indeed they assume that the bias is matrix-monotone, which is not true in general.

For kernel ridge regression, the bias of the estimator $\hat{f}_K$ can be expressed as
\begin{align*}
\text{bias}(K)^2 &= n\lambda^2 \|(K+n\lambda I)^{-1}f^*\|^2\\
&= n\lambda^2 {f^*}^\top(K+n\lambda I)^{-2}f^*. 
\end{align*}

For $\gamma >0$, we consider again the regularized approximation $L_{\gamma} = KS(S^\top KS + n\gamma I)^{-1}S^\top K$ with $S\in\R^{n\times p}$ the sketching matrix. The result of the theorem follows from the three following lemmas.
 
 \begin{lemma}\label{lemm1}
Let $K = U\Sigma U^\top$ where $U$ is orthogonal and $\Sigma$ diagonal positive. We have 
\begin{equation}
L_{\gamma} \preceq L \preceq K .
\label{order}
\end{equation}
Moreover, let \[D =  \Phi - \Phi^{1/2}U^\top SS^\top U\Phi^{1/2}\] with $\Phi = \Sigma(\Sigma +n\gamma I)^{-1}$. If $\lambda_{\max}(D)\leq t$ for $t \in (0,1)$ then
\[0 \preceq K - L_{\gamma} \preceq \frac{n\gamma}{1-t} I.\]
\label{basic_ineq}
\end{lemma}
\begin{lemma}\label{lemm2}
If $0 \preceq K - L_{\gamma} \preceq \frac{n\gamma}{1-t} I$ then $\text{bias}(L_\gamma) \leq  \left(1+\frac{\gamma/\lambda}{1-t}\right)\text{bias}(K)$.
\end{lemma}
%If one is interested in the regularized Nystr\"{o}m approximation $L_\gamma = KS(S^\top KS+n\gamma I)^{-1}S^\top K$ instead of $L = KS(S^\top KS)^{\dagger}S^\top K$, then the following lemma is not needed. 
\begin{lemma}\label{lemm3}
If $0 \preceq K - L_{\gamma} \preceq \frac{n\gamma}{1-t} I$ and $\lambda \ge \frac{1}{1-t}\|S\|_{\text{op}}^2 \cdot\frac{\lambda_{\max}(K)}{n}$ then the map $\gamma \rightarrow \text{bias}(L_\gamma)$ is increasing. This in particular implies that under the same conditions, $\text{bias}(L) \le \text{bias}(L_\gamma)$.%$\lambda > \frac{1}{n}\|S^\top K^2 S (S^\top K S)^{\dagger}\|_{\text{op}}$
\end{lemma}
\noindent We next prove the above lemmas.

\noindent\begin{proofof}{Lemma \ref{lemm1}}
With $K = U\Sigma U^\top$ and $R = \Sigma^{1/2}U^\top S$, $\bar{L}_\gamma = R (R^\top R + n\gamma I)^{-1} R^\top$, we have
\[L_{\gamma} = U\Sigma^{1/2} \bar{L}_\gamma \Sigma^{1/2}  U^\top.\]
 Due to the matrix inversion lemma, we have 
\begin{align*} 
\bar{L}_{\gamma} &= RR^\top(RR^\top + n\gamma I)^{-1} \\
&= I - n\gamma (RR^\top + n\gamma I)^{-1}\\
&= I - n\gamma ( \Sigma + n\gamma I +  RR^\top - \Sigma)^{-1}\\
&= I - n\gamma (\Sigma + n\gamma I)^{-1/2} (  I -  D )^{-1}(\Sigma + n\gamma I)^{-1/2} 
\end{align*}
with 
\begin{align*}
D &= (\Sigma + n\gamma I)^{-1/2}(\Sigma - RR^\top)(\Sigma + n\gamma I)^{-1/2} \\
&= \Phi - \Phi^{1/2}U^\top SS^\top U\Phi^{1/2},
\end{align*}
and $\Phi = \Sigma(\Sigma +n\gamma I)^{-1}$. This shows that for any $\gamma \geq 0$ 
\[L_{\gamma} \preceq L \preceq K .\]

Now if $\lambda_{\max}(D)\leq t$ for $t \in (0,1)$, 
\[I - \bar{L}_\gamma \preceq \frac{n\gamma}{1-t} (\Sigma +n\gamma I)^{-1} \]
which implies 

\[0 \preceq K - L_{\gamma} \preceq \frac{n\gamma}{1-t}K(K+n\gamma I)^{-1} \preceq \frac{n\gamma}{1-t} I.\]
\end{proofof}
%Then, assuming $\frac{\gamma/\lambda}{1-t} \leq 1$, the previous inequality implies
%\begin{align*} 
%\big(L_\gamma+n\lambda I\big)^{-1} &\preceq \big(K - \frac{n\gamma}{1-t} I + n\lambda I\big)^{-1} \\
%&= \big(K + n\lambda(1-\frac{\gamma/\lambda}{1-t}) I\big)^{-1} \\
%&\preceq (1-\frac{\gamma/\lambda}{1-t})^{-1}\big(K+n\lambda I\big)^{-1}. 
%\end{align*} 
%Hence \[\text{bias}(L_\gamma)^2 \leq  \Big(1-\frac{\gamma/\lambda}{1-t}\Big)^{-2}\text{bias}(K)^2.\]
%
%From $L_{\gamma} \preceq L$, and the fact that $\text{bias}(K)$ is matrix-decreasing in $K$, we obtain the desired result.

\noindent\begin{proofof}{Lemma \ref{lemm2}} This proof was communicated to us by Francis Bach \cite{personal}.

\noindent Since $K-L_{\gamma}$ commutes with the identity,
we have 
\[(K-L_{\gamma})^2 \preceq \frac{n^2\gamma^2}{(1-t)^2} I.\] 
Now,
\begin{align*}
\|(L_{\gamma}+n\lambda I)^{-1}f^* - (K+n\lambda I)^{-1}f^*\|_2 &= \|(L_{\gamma}+n\lambda I)^{-1}(K - L_{\gamma})(K+n\lambda I)^{-1}f^*\|_2\\
&\le \|(L_{\gamma}+n\lambda I)^{-1}(K - L_{\gamma})\|_{\text{op}} \cdot \|(K+n\lambda I)^{-1}f^*\|_2.
\end{align*}
On the other hand,
\begin{align*}
\|(L_{\gamma}+n\lambda I)^{-1}(K - L_{\gamma})\|_{\text{op}}^2 &= \|(L_{\gamma}+n\lambda I)^{-1}(K - L_{\gamma})^2(L_{\gamma}+n\lambda I)^{-1}\|_{\text{op}} \\
&\le \frac{n^2\gamma^2}{(1-t)^2} \|(L_{\gamma}+n\lambda I)^{-2}\|_{\text{op}}\\
&\le \frac{n^2\gamma^2}{(1-t)^2} \|(L_{\gamma}+n\lambda I)^{-1}\|_{\text{op}}^2.
\end{align*}
This yields,
\begin{align*}
\|(L_{\gamma}+n\lambda I)^{-1}f^*\|_2 &\le \|(K+n\lambda I)^{-1}f^*\|_2 + \|(L_{\gamma}+n\lambda I)^{-1}f^* - (K+n\lambda I)^{-1}f^*\|_2\\
&\le  \|(K+n\lambda I)^{-1}f^*\|_2 \cdot \left(1+\frac{n\gamma}{1-t}\|(L_{\gamma}+n\lambda I)^{-1}\|_{\text{op}}\right)\\
&\le \|(K+n\lambda I)^{-1}f^*\|_2 \cdot \left(1+\frac{\gamma/\lambda}{1-t}\right).
\end{align*}
Hence we have the bias inequality
\[\text{bias}(L_\gamma) \leq  \left(1+\frac{\gamma/\lambda}{1-t}\right)\text{bias}(K).\] 
\end{proofof}

\noindent\begin{proofof}{Lemma \ref{lemm3}}
Let $\varphi(\gamma) = {f^*}^\top(L_\gamma+n\lambda I)^{-2}f^*$. The task is to prove that $\varphi$ is increasing if $\lambda \ge \frac{1}{1-t}\|S\|_{\text{op}}^2 \frac{\lambda_{\max}(K)}{n}$. We do so by computing the derivative of $\varphi$ and showing that $\varphi' \ge 0$. Let $\gamma,\gamma' >0$. We have
\begin{align*}
\varphi(\gamma) - \varphi(\gamma') &= {f^*}^\top\left((L_\gamma+n\lambda I)^{-2}-(L_{\gamma'}+n\lambda I)^{-2}\right)f^*\\
&= {f^*}^\top(L_\gamma+n\lambda I)^{-2}\left((L_{\gamma'}+n\lambda I)^{2}-(L_\gamma+n\lambda I)^{2}\right)(L_{\gamma'}+n\lambda I)^{-2}f^*\\
&= {f^*}^\top(L_\gamma+n\lambda I)^{-2}\left((L_{\gamma'}^{2}-L_\gamma^{2}) +2n\lambda (L_{\gamma'}-L_\gamma)\right)(L_{\gamma'}+n\lambda I)^{-2}f^*.
\end{align*}
Now we compute the terms $L_{\gamma'}-L_\gamma$ and $L_{\gamma'}^2-L_\gamma^2$:
\begin{align*}
L_{\gamma'}-L_\gamma &= KS(S^\top K S + n\gamma' I)^{-1}S^\top K - KS(S^\top K S + n\gamma I)^{-1}S^\top K \\
&=KS (S^\top K S + n\gamma' I)^{-1} \left(n (\gamma - \gamma')\right) (S^\top K S + n\gamma I)^{-1} S^\top K.
\end{align*}
And
\begin{align*}
L_{\gamma'}^2-L_\gamma^2 &= KS(S^\top K S + n\gamma' I)^{-1}S^\top K^2 S(S^\top K S + n\gamma' I)^{-1}S^\top K \\
& ~~~~- KS(S^\top K S + n\gamma I)^{-1}S^\top K^2 S(S^\top K S + n\gamma I)^{-1}S^\top K\\
&=KS(S^\top K S + n\gamma' I)^{-1}S^\top K^2 S(S^\top K S + n\gamma' I)^{-1}S^\top K\\
&~~~~ - KS(S^\top K S + n\gamma' I)^{-1}S^\top K^2 S(S^\top K S + n\gamma I)^{-1}S^\top K\\
&~~~~+KS(S^\top K S + n\gamma' I)^{-1}S^\top K^2 S(S^\top K S + n\gamma I)^{-1}S^\top K\\
& ~~~~- KS(S^\top K S + n\gamma I)^{-1}S^\top K^2 S(S^\top K S + n\gamma I)^{-1}S^\top K\\
&= KS(S^\top K S + n\gamma' I)^{-1}S^\top K^2 S \left[(S^\top K S + n\gamma' I)^{-1} - (S^\top K S + n\gamma I)^{-1}\right]S^\top K \\
&~~~~+ KS\left[(S^\top K S + n\gamma' I)^{-1} - (S^\top K S + n\gamma I)^{-1}\right]S^\top K^2 S (S^\top K S + n\gamma' I)^{-1} S^\top K.
\end{align*}
The first term is the last equality above is equal to 
\[ n (\gamma - \gamma') \cdot KS(S^\top K S + n\gamma' I)^{-1}S^\top K^2 S (S^\top K S + n\gamma' I)^{-1} (S^\top K S + n\gamma I)^{-1}S^\top K,\]
and the second one is equal to
\[ n (\gamma - \gamma') \cdot KS(S^\top K S + n\gamma' I)^{-1}(S^\top K S + n\gamma I)^{-1}S^\top K^2 S (S^\top K S + n\gamma' I)^{-1} S^\top K.\] 
 Now combining the above and taking the limit $\gamma' \rightarrow \gamma$ we have 
 \begin{align*}
\lim_{\gamma' \rightarrow \gamma} &\frac{\varphi(\gamma) - \varphi(\gamma')}{n(\gamma - \gamma')} =\\
  &{f^*}^\top(L_\gamma+n\lambda I)^{-2}KS(S^\top K S + n\gamma I)^{-1} \cdot Q \cdot (S^\top K S + n\gamma I)^{-1} S^\top K (L_{\gamma}+n\lambda I)^{-2}f^*,
 \end{align*}
 with 
 \[Q = 2n\lambda I + S^\top K^2 S (S^\top K S + n\gamma I)^{-1} + (S^\top K S + n\gamma I)^{-1}S^\top K^2 S : = 2n\lambda I+ \bar{Q}.\]
 
Therefore, the function $\varphi$ is increasing for all $\gamma$ such that $Q \succeq 0$, and the latter is true if $2n\lambda \ge -\lambda_{\min}(\bar{Q})$. Moreover, since $\bar{Q}$ is symmetric we have
\[\lambda_{\min}(\bar{Q}) \ge -\|\bar{Q}\|_{\text{op}} \ge -2\|S^\top K^2 S (S^\top K S + n\gamma I)^{-1}\|_{\text{op}},\] 
and it is sufficient to verify the condition
\begin{equation} \label{cond_lambda}
n \lambda \ge \|S^\top K^2 S (S^\top K S + n\gamma I)^{-1}\|_{\text{op}}.
\end{equation} 

Now we finish the proof by showing that the above operator norm is smaller than $\frac{1}{1-t}\|S\|_{\text{op}}^2 \lambda_{\max}(K)$. We have
\begin{align*}
n\gamma S^\top K^2 S (S^\top K S + n\gamma I)^{-1} &= S^\top K^2 S (S^\top K S + n\gamma I)^{-1}(n\gamma I + S^\top K S - S^\top K S )\\
&= S^\top K^2 S - S^\top K^2 S (S^\top K S + n\gamma I)^{-1} S^\top K S \\
&= S^\top K( K -  K S (S^\top K S + n\gamma I)^{-1} S^\top K) S \\
&= S^\top K (K-L_\gamma) S.
\end{align*}
Taking operator norms, and using the assumption $0\preceq K - L_{\gamma} \preceq \frac{n\gamma}{1-t} I$,
\[n\gamma \|S^\top K^2 S (S^\top K S + n\gamma I)^{-1}\|_{\text{op}} \le \|S^\top\|_{\text{op}} ~ \|K\|_{\text{op}}~\frac{n\gamma}{1-t} ~\|S\|_{\text{op}}.\]
Hence, \eqref{cond_lambda} is satisfied if $n\lambda \ge \frac{1}{1-t}\|S\|_{\text{op}}^2\lambda_{\max}(K)$ therefore concluding the proof.
\end{proofof}

%\newpage
%% MWM: The following line should be commented out if the appendix is in a different file that the main text
\section{Proof of Theorem~\ref{concent}}
%% MWM: The following line should be commented out if the appendix is in the same file as the main text
%\section{Proof of Theorem 2}
\label{sxn:pf-concent}

The proof uses the matrix Bernstein inequality (see e.g. Theorem 6.1.1 in \cite{tropp2012user}):
\begin{thm} \label{bernstein1}
Consider a sequence $(X_k)$ of independent random symmetric matrices with dimension $d$. Assume that $\ex(X_k)=0$, $\lambda_{\max}(X_k) \leq R$, and let $Y = \sum_k X_k$. Furthermore, assume that there exists $\sigma >0$ such that $\|\ex(Y^2)\|_2 \leq \sigma^2$. Then \[\prob\big(\lambda_{\max}(Y) \geq t\big) \leq d\exp\Big(\frac{-t^2/2}{\sigma^2 + Rt/3}\Big).\]
\end{thm}
Next , we exhibit the sequence $(X_k)$ and $Y$ in our case. We have \[\Psi\Psi^\top = \sum_{i=1}^m \psi_i\psi^\top_i\]  and\[ \Psi SS^\top \Psi^\top = \frac{1}{p}\sum_{i\in I} \frac{1}{p_i}\psi_i\psi_i^\top = \frac{1}{p}\sum_{i=1}^m \sum_{k=1}^p\frac{1}{p_i}z_{ik}\psi_i\psi_i^\top\]
where $(z_{ik})_{1\le i\le m}$ are i.i.d.\ binary random vectors for $k \in \{1,\cdots,p\}$ with $\prob(z_{ik} = 1) =p_i$ (i.e.\ $(z_{ik})_{1\le i\le m}$ is the indicator of the chosen column at trial $k$). Let $Y = \Psi \Psi^\top - \Psi SS^\top\Psi^\top$, then \[Y = \frac{1}{p} \sum_{k=1}^p \sum_{i=1}^m(1-\frac{z_{ik}}{p_i})\psi_i\psi_i^\top.\] 
We choose $X_k$ to be $\frac{1}{p}\sum_{i=1}^m(1-\frac{z_{ik}}{p_i})\psi_i\psi_i^\top$ for every $k \in \{1,\cdots,p\}$. Now we verify the assumptions of the above theorem. The matrices $X_k$ inherit independence from the random vectors $(z_{ik})_{1\le i \le m}$, and we have $\ex(X_k) = 0$, and $\lambda_{\max}(X_k) \leq \frac{1}{p}\lambda_{\max}(\sum_{i=1}^m\psi_i\psi_i^\top) = \frac{1}{p}\lambda_{\max}(\Psi\Psi^\top)$. Now we control the spectral norm of the second moment of $Y$. Again with $\ex(X_k)=0$ we have
$\ex(Y^2) = \sum_{k,k'=1}^p\ex(X_kX_{k'}) = \sum_{k=1}^p\ex(X_k^2).$ 
And for $k \in \{1,\cdots,p\}$
\begin{align*}
\ex(X_k^2) &= \frac{1}{p^2}\sum_{i,i'=1}^m \ex\left(\left(1-\frac{z_{ik}}{p_i}\right)\left(1-\frac{z_{i'k}}{p_{i'}}\right)\right)\psi_{i'}\psi_{i'}^\top\psi_i\psi_i^\top \\
&= \frac{1}{p^2}\sum_{i,i'=1}^m \left(\frac{\ex(z_{ik}z_{i'k})}{p_ip_{i'}} - 1\right)\psi_{i'}\psi_{i'}^\top\psi_i\psi_i^\top. 
\end{align*}
To proceed, observe that for $i \neq i'$, $z_{ik}z_{i'k} = 0$ since only one column is chosen at a time. This yields
\begin{align*}
\ex(X_k^2) &=  \frac{1}{p^2}\sum_{i=1}^m \frac{\ex(z_{ik}^2)}{p_i^2}\psi_i\psi_i^\top\psi_i\psi_i^\top  - \frac{1}{p^2}\sum_{i,i'=1}^m \psi_{i'}\psi_{i'}^\top\psi_i\psi_i^\top\\
&=  \frac{1}{p^2}\sum_{i=1}^m \frac{1}{p_i}\|\psi_i\|_2^2\psi_i\psi_i^\top - \left(\frac{1}{p}\sum_{i=1}^m\psi_i\psi_i^\top\right)^{2} \\
&\preceq \frac{1}{p^2}\sum_{i=1}^m \frac{\|\psi_i\|_2^2}{p_i}\psi_i\psi_i^\top.
\end{align*}
Given that the probability distribution $(p_i)$ verifies $p_i \geq \beta \frac{\|\psi_i\|_2^2}{\|\Psi\|_F^2}$, we get 
$\ex(Y^2) \preceq \frac{\|\Psi\|_F^2}{\beta p}\sum_{i=1}^m\psi_i\psi_i^\top = \frac{\|\Psi\|_F^2}{\beta p}\Psi\Psi^\top.$
Hence $\|\ex(Y^2)\|_2 \leq \frac{\|\Psi\|_F^2}{\beta p}\lambda_{\max}(\Psi\Psi^\top). $
We now apply the theorem with $R = \frac{1}{p}\lambda_{\max}(\Psi\Psi^\top)$ and $\sigma^2 = \frac{\|\Psi\|_F^2}{\beta p}\lambda_{\max}(\Psi\Psi^\top)$ which leads to the desired result.

%% MWM: The following line should be commented out if the appendix is in a different file that the main text
\section{Proof of Theorem~\ref{main}}
%% MWM: The following line should be commented out if the appendix is in the same file as the main text
%\section{Proof of Theorem 3}
\label{sxn:pf-main}

\paragraph{Monotonicity of the variance.} First of all, we observe that the variance of the estimator $\hat{f}_K$ is matrix-increasing as a function of $K$. Indeed, we have
\[\text{variance}(K) = \frac{\sigma^2}{n}\trace(K^2(K+n\lambda I)^{-2}) =  \frac{\sigma^2}{n}\sum_{j=1}^n\frac{\lambda_j(K)^2}{(\lambda_j(K)+n\lambda)^2},\]  
where $\lambda_j(K)$ is the $j$th eigenvalue of $K$ arranged in a decreasing order. The function $x \rightarrow \frac{x^2}{(x+n\lambda)^2}$ is increasing for $x \ge 0$. Moreover, if $L  \preceq K$ then by the Courant-Fischer minimax principle $\lambda_j(L) \le \lambda_j(K)$ for all $j$ (e.g.\ see Corollary III.1.2 in \cite{bhatia2013matrix}). 
 
\paragraph{Risk bound.} Now, using Theorem \ref{bias_} combined with the above fact, we have
\begin{align*}
\ex_{\xi} \|\hat{f}_L - f^*\|_2^2  &= \text{bias}(L)^2 + \text{variance}(L) \\
&\leq \left(1+ \frac{\gamma/\lambda}{1-t}\right)^{2}\text{bias}(K)^2 + \text{variance}(K) \\
&\leq \left(1+\frac{\gamma/\lambda}{1-t}\right)^{2}(\text{bias}(K)^2 + \text{variance}(K)) \\
&= \left(1+ \frac{\gamma/\lambda}{1-t}\right)^{2}\ex_{\xi} \|\hat{f}_L - f^*\|_2^2
\end{align*}%(which verifies $\gamma \leq (1-t) \lambda$)
We set $\gamma = \lambda\epsilon$ and $t=1/2$. The above holds if $\lambda_{\max}\Big(\Phi - \Phi^{1/2} U^\top SS^\top U\Phi^{1/2} \Big)\leq t$ and $n\lambda \ge \frac{1}{1-t}\|S\|_{\text{op}}^2\lambda_{\max}(K)$. Now let $\Psi = \Phi^{1/2}U^\top$. Then we have $\|\psi_i\|_2^2 = l_i(\gamma)$ and $\|\Psi\|_F^2 = d_{\text{eff}}$. Using Theorem \ref{concent} on $\Psi$, and given that $\lambda_{\max}(\Psi\Psi^\top) =\lambda_{\max}(\Phi) \leq 1$, for the result to hold with probability at least $1-\rho$, it is sufficient to set $p$ such that $n\exp\Big(\frac{-p(1/2)^2/2}{d_{\text{eff}}/\beta +1/6}\Big) \leq \rho$ which gives the desired lower bound $p \geq 8(d_{\text{eff}}/\beta + 1/6)\log\left(\frac{n}{\rho}\right)$. 

\noindent\textbf{Remark:} Note that if one uses the regularized Nystr\"{o}m approximation $L_\gamma = KS(S^\top KS+n\gamma I)^{-1}S^\top K$ with $\gamma =\lambda\epsilon$ instead of $L = KS(S^\top KS)^{\dagger}S^\top K$ in the algorithm then the proof would now be complete and the condition condition $n\lambda \ge \frac{1}{1-t}\|S\|_{\text{op}}^2\lambda_{\max}(K)$ is not necessary. If one uses $L$, then this latter condition needs to be verified to insure monotonicity of the bias (see Lemma \ref{lemm3}). 

\paragraph{Controlling $\|S\|_{\text{op}}$.} Now it remains to control the operator norm of the sketching matrix $S$ appearing in the lower bound on $\lambda$.  To this end we use a variant of the matrix Bernstein inequality (Theorem \ref{bernstein1}) for controlling operator norms of random matrices (see Corollary 6.2.1 in \cite{tropp2012user}).

\begin{thm} Consider a sequence $(X_k)$ of independent random symmetric matrices with dimension $d \times d$. Assume that $\ex(X_k)=0$, $\|X_k\|_{\text{op}} \leq R$, and let $Y = \sum_k X_k$. Furthermore, assume that there exists $\sigma >0$ such that $\|\ex(Y^2)\|_{\text{op}} \leq \sigma^2$. Then \[\prob\left(\|Y\|_{\text{op}} \geq t \right) \leq 2d\exp\left(\frac{-t^2/2}{\sigma^2 + Rt/3}\right).\]
\label{bernstein2}
\end{thm}
 We are interested in the sum
\[Y = SS^\top-I = \frac{1}{p} \sum_{k=1}^p \sum_{i=1}^n\left(\frac{z_{ik}}{p_i}-1\right)e_ie_i^\top,\]
and similarly to the previous section we consider the sequence $X_k = \frac{1}{p}\sum_{i=1}^n(\frac{z_{ik}}{p_i}-1)e_ie_i^\top$ where $z_{ik}$ is defined as before and $(e_i)_{1\le i \le n}$ in the standard basis in $\R^n$.
Since $p_i \ge \beta \cdot l_i(\lambda\epsilon)/d_{\text{eff}}$ with $d_{\text{eff}} = \sum_{i=1}^n l_i(\lambda\epsilon)$ we have 
\[\|X_k\|_{\text{op}} \le \frac{1}{p} \max_i \left(\frac{d_{\text{eff}}}{\beta l_i(\lambda\epsilon)} - 1\right) = \frac{1}{p} \left(\frac{d_{\text{eff}}}{\beta \underline{l}} - 1\right) \le \frac{d_{\text{eff}}}{p\beta \underline{l}},\]%for any given $k$, among all the $z_{ik}$'s, $1\le i\le n$ only one of them is equal to 1, and
with $\underline{l} = \min_i l_i(\lambda\epsilon)$. On the other hand,
\[\ex(X_k^2) = \frac{1}{p^2}\sum_{i=1}^n\ex\left(\left(\frac{z_{ik}}{p_i}-1\right)^2\right) e_ie_i^\top = \frac{1}{p^2}\sum_{i=1}^n\left(\frac{1}{p_i}-1\right) e_ie_i^\top \preceq \frac{1}{p^2} \frac{d_{\text{eff}}}{\beta \underline{l}}I.\]
Hence 
\[\|\ex(Y^2)\|_{\text{op}} \le \frac{1}{p}\frac{d_{\text{eff}}}{\beta \underline{l}}.\]
By choosing $\sigma^2 = R = \frac{1}{p} \frac{d_{\text{eff}}}{\beta \underline{l}}$, we have $\|SS^\top-I\|_{\text{op}} \le t$ with probability at least $1-2n\exp\left(-\frac{t^2/2}{R(1+t/3)}\right)$. Taking $t = \max\left\{1~,~\frac{8d_{\text{eff}}}{3\beta \underline{l}\cdot p}\log \left(\frac{2n}{\rho}\right)\right\}$, the latter probability is greater than $1-\rho$, and by the triangle inequality: $\|S\|_{\text{op}}^2 \le 1 + t$ with the same probability. By taking $p \ge 8(d_{\text{eff}}/\beta + 1/6)\log\left(\frac{n}{\rho}\right)$ (thereby verifying the condition from the previous paragraph) we have
\begin{align*}
\frac{8d_{\text{eff}}}{3\beta \underline{l}\cdot p}\log \left(\frac{2n}{\rho}\right) \le \frac{1}{3 \underline{l}} \cdot \frac{d_{\text{eff}}}{(d_{\text{eff}} + \beta/6)} \cdot \frac{ \log \left(\frac{2n}{\rho}\right)}{ \log\left(\frac{n}{\rho}\right)} 
 \le \frac{1}{3 \underline{l}}  \cdot \left(1 + \frac{ \log 2}{ \log\left(\frac{n}{\rho}\right)}\right) \le \frac{1}{\underline{l}}
\end{align*}
if $n \ge 2$, and therefore $\|S\|_{\text{op}}^2 \le 1+1/\underline{l}$ (since $\underline{l} \le 1$) with probability at least $1-\rho$. 

%% MWM: The following line should be commented out if the appendix is in a different file that the main text
\section{Proof of Theorem~\ref{approx_lev}}
%% MWM: The following line should be commented out if the appendix is in the same file as the main text
%\section{Proof of Theorem 4}
\label{sxn:pf-approx_lev}

First, it is clear that 
\begin{align*} \tilde{l}_i &= e_i^\top B(B^\top B+n\lambda I)^{-1}B^{\top}e_i \\
&= e_i^\top BB^\top(BB^\top+n\lambda I)^{-1}e_i  \\
&= \text{diag}(L(L+n\lambda I)^{-1})_{i}
\end{align*}
with $e_i$ the $i$-th element of the standard basis in $\R^n$. Now we bound the approximations $\tilde{l}_i$ by comparing the matrices $L(L+n\lambda I)^{-1}$ and $K(K+n\lambda I)^{-1}$ with respect to the semidefinite order. Since $L \preceq K$ (Appendix A) and the map $K \rightarrow K(K+n\lambda I)^{-1}$ is matrix-increasing, we immediately get the upper bound $\tilde{l}_i \leq l_i(\lambda)$ for all $i \in \{1,\cdots,n\}$. Next we derive the lower bound. %\eqref{order} see \eqref{order}
For $\gamma >0$, we consider again the regularized approximation $L_{\gamma} = KS(S^\top KS + n\gamma I)^{-1}S^\top K$ with $S\in\R^{n\times p}$ the sketching matrix. Due the matrix inversion lemma, $L_{\gamma} \preceq L$ (Appendix A). Hence to get a lower bound on $\tilde{l}_i$, it suffices to obtain a lower bound for the same quantity when $L$ is replaced by $L_{\gamma}$. We proved in Appendix A that if \[\lambda_{\max}\Big(\Psi\Psi^\top - \Psi SS^\top \Psi^\top \Big)\leq t\] for $t\geq 0$ with $\Psi =\Phi^{1/2}U^\top$, $\Phi = \Sigma(\Sigma +n\gamma I)^{-1}$ then \[K - L_{\gamma} \preceq \frac{n\gamma}{1-t}K(K+n\gamma I)^{-1}\preceq \frac{n\gamma}{1-t}I.\] Therefore 
\begin{align*} L_{\gamma}(L_{\gamma}+n\lambda I)^{-1}  &\succeq (K-\frac{n\gamma}{1-t}I)(K+n\lambda I)^{-1}\\
&\succeq K(K+n\lambda I)^{-1} - \frac{\gamma/\lambda}{1-t}I,
\end{align*}
where the last line follows by distributing the product and using the inequality $K + n\lambda I \succeq n\lambda I$ for the second term. Hence $\tilde{l}_i \geq l_i(\lambda) - \frac{\gamma/\lambda}{1-t}$. Now we choose again $t = 1/2$ and $\gamma = \epsilon \lambda$ for $\epsilon \in (0,1/2)$, we get the additive error bound on $\tilde{l}_i$ and similarly to the proof of Theorem \ref{main}, it suffices to have $p \geq 8(d_{\text{eff}}/\beta + 1/6)\log\left(\frac{n}{\rho}\right)$. 
To finish the proof, we choose the sampling distribution $(p_i)_i$ and $\beta$ appropriately. Since 
\[l_i(\gamma)  = \sum_{j=1}^n \frac{\sigma_j}{\sigma_j + n\gamma}U_{ij}^2 \leq \sum_{j=1}^n \frac{\sigma_j}{n\gamma}U_{ij}^2 = \frac{1}{n\gamma}K_{ii},\] 
by choosing $p_i = K_{ii}/\trace(K)$, we have $p_i \geq \beta ~ l_i(\lambda\epsilon)/ \sum_{i=1}^n l_i(\lambda\epsilon)$ with $\beta = n\lambda\epsilon d_{\text{eff}}/\trace(K)$, which yields $d_{\text{eff}}/\beta = \trace(K)/(n\lambda\epsilon)$. 

\noindent As for the multiplicative error bound, using $K - L_{\gamma} \preceq \frac{n\gamma}{1-t}K(K+n\gamma)^{-1}$  we get 
\begin{align*}
 L_{\gamma}(L_{\gamma} & +n\lambda I)^{-1} \succeq (K-\frac{n\gamma}{1-t}K(K+n\gamma)^{-1})(K+n\lambda I)^{-1} \\
&= K(K+n\lambda I)^{-1}(I - \frac{n\gamma}{1-t}(K+n\gamma I)^{-1}).
\end{align*}
For $t=1/2$, $I - \frac{n\gamma}{1-t}(K+n\gamma I)^{-1} = (K-n\gamma I)(K+n\gamma I)^{-1} \succeq \frac{\sigma_{n}-n\gamma}{\sigma_{n}+n\gamma} I$. The result follows.

\end{document}